\documentclass{article}
\usepackage{cite}
\usepackage{stfloats}
\usepackage{spconf,amsmath,graphicx}
\usepackage{subfiles}
\usepackage{tabularx}
\usepackage[symbol]{footmisc}
\usepackage{amsmath,amssymb,amsfonts}
\usepackage{algorithmic}
\usepackage{graphicx}
\usepackage{textcomp}
\usepackage{xcolor}
\usepackage{url}
\usepackage{caption}

\title{Ridgeformer: Mutli-Stage Contrastive Training For Fine-grained Cross-Domain Fingerprint Recognition}
\name{Shubham Pandey$\ast$\thanks{$\ast$ equal contribution first authors}, Bhavin Jawade$\ast$, Srirangaraj Setlur}
\address{University at Buffalo, The State University of New York}
\begin{document}
%
\maketitle

\begin{figure*}[!b]
  \noindent
  \begin{minipage}{\textwidth}
    \vspace{0.5em}    
    \footnotesize
    © 2025 IEEE. Personal use of this material is permitted. Permission from IEEE must be obtained for all other uses, in any current or future media, including reprinting/republishing this material for advertising or promotional purposes, creating new collective works, for resale or redistribution to servers or lists, or reuse of any copyrighted component of this work in other works.
  \end{minipage}
\end{figure*}

\begin{abstract}
The increasing demand for hygienic and portable biometric systems has underscored the critical need for advancements in contactless fingerprint recognition. Despite its potential, this technology faces notable challenges, including out-of-focus image acquisition, reduced contrast between fingerprint ridges and valleys, variations in finger positioning, and perspective distortion. These factors significantly hinder the accuracy and reliability of contactless fingerprint matching. To address these issues, we propose a novel multi-stage transformer-based contactless fingerprint matching approach that first captures global spatial features and subsequently refines localized feature alignment across fingerprint samples. By employing a hierarchical feature extraction and matching pipeline, our method ensures fine-grained, cross-sample alignment while maintaining the robustness of global feature representation. We perform extensive evaluations on publicly available datasets such as HKPolyU and RidgeBase under different evaluation protocols, such as contactless-to-contact matching and contactless-to-contactless matching and demonstrate that our proposed approach outperforms existing methods, including COTS solutions. Our codebase is available at \url{https://github.com/KNITPhoenix/Ridgeformer}
\end{abstract}
\section{Introduction}
\label{sec:intro}

\begin{figure}[t!]
\centering
     \includegraphics[scale=0.4]{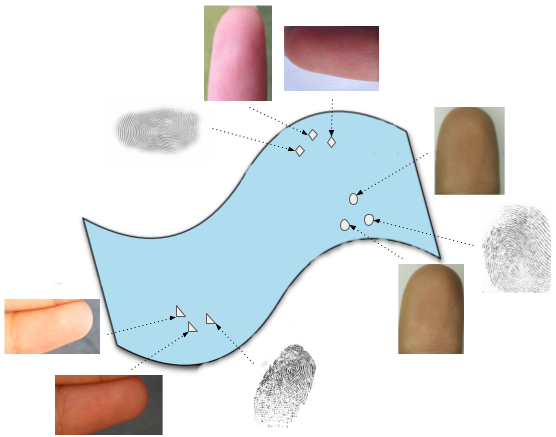}
    \caption{An illustration demonstrating the objective of bringing cross domain images from same subject close to each other increasing the similarity, while pushing the images from different subjects far from each other.}
    \label{fig:verticalcell}
\vspace{-1.5em}
\end{figure}

The demand for hygienic, portable, and robust biometric systems continues to grow, particularly in applications requiring secure, touchless authentication. While traditional contact-based fingerprint recognition remains dominant, it faces several challenges, including latent fingerprint theft, hygiene concerns, and the complexity of deployment in uncontrolled or remote environments. These drawbacks have driven interest in contactless fingerprint recognition, which eliminates physical contact, addressing hygiene risks and enhancing accessibility.

Despite the promise of contactless fingerprint recognition, it encounters significant obstacles such as out-of-focus image acquisition, reduced ridge-valley contrast, finger-angle variations, and perspective distortions. These factors complicate accurate fingerprint matching, particularly when comparing contactless and contact-based fingerprints. Bridging the gap between these two types of fingerprint images is crucial for enabling seamless integration into biometric systems.

In this work, we propose a novel approach for contactless-to-contact fingerprint matching, addressing the domain shift between contactless and contact-based images. Our method learns a unified latent space for both fingerprint types, allowing for more effective cross-domain matching. The core of our approach is a multi-stage architecture that first captures global spatial features using a Vision Transformer (ViT) \cite{dosovitskiy2020image}, followed by a fine-grained local alignment stage to account for detailed fingerprint characteristics that may be lost in global representations. This hierarchical feature extraction pipeline enhances robustness and accuracy in matching contactless and contact-based fingerprints.

Our approach is evaluated on the publicly available datasets HKPolyU \cite{polyudataset} and RidgeBase \cite{jawade2022ridgebase} under different matching protocols, including contactless-to-contact and contactless-to-contactless matching, where it demonstrates superior performance compared to existing methods, including COTS solutions.

The key contributions of this work are:
\begin{enumerate}
    \setlength{\itemsep}{0pt}
    \item To the best of our knowledge, this is the first work on contactless-to-contact fingerprint matching that employs a vision transformer based architecture.
    \item We propose an multi-stage training strategy that utilizes intra-sample cross-attention to compute fine-grained alignment score between fingerprints.
    \item Experimental evaluations on the HKPolyU \cite{polyudataset} and RidgeBase \cite{jawade2022ridgebase} datasets, demonstrating superior performance in the contactless-to-contact and contactless-to-contactless fingerprint matching scenarios.
\end{enumerate}

\vspace{-1.0em}
\section{Related Works}
\label{sec:format}
\noindent In this section, we will review relevant contributions in multiple areas that have laid the groundwork for our proposed method. \\
\textbf{Contactless Matching} With a recent shift towards contactless fingerprint recognition, numerous datasets have been developed \cite{grosz2021c2cl, ispfdv2, polyudataset, jawade2022ridgebase}. The RidgeBase \cite{jawade2022ridgebase} dataset, consisting of 15,000 contactless and contact-based fingerprints collected from 88 individuals, was designed for single-finger and multi-finger matching for CL2CL and CL2CB verification and identification. In addition, there has also been research done in the area of synthetic fingerprint generation \cite{priesnitz2022syncolfinger, bouzaglo2022synthesis, long20153d, maltoni2009synthetic, dong2023synthesis}. During data collection, it can be observed that improper finger positioning when using a contactless fingerprint sensor can lead to distortions and deformations. To this end, Grosz et al \cite{grosz2021c2cl} have proposed an end-to-end system with preprocessing and matching algorithms. Tan et al \cite{tan2020towards} have developed a framework that eliminates the need for image enhancement while \cite{labati2013contactless} propose a CNN-based approach.  Differences in the images captured by each modality’s sensors pose significant problems in CL2CB matching. To solve this, a novel minutiae attention network proposed by Tan et al \cite{tan2021minutiae} uses a Siamese network with reciprocal distance loss to learn robust global and local minutiae features for identification. Lin et al \cite{polyudataset} have proposed an approach using robust thin-plate spline (RTPS) to correct deformations and distortions and ensure the correct alignment of key minutiae features from both modalities. Lin et al \cite{lin2018cnn} have also designed a system using a multi-Siamese network to learn deep fingerprint representations, another significant challenge in contactless matching. Grosz et al \cite{grosz2022minutiae} have proposed the use of a Vision Transformer (ViT) to learn fingerprint embeddings of a fixed length using minutiae features. MRA-GNN, designed by Su et al \cite{su2023mra}, utilizes a GNN to learn descriptive features based on the topology and correlation of fingerprints. \\
\textbf{Deep Metric Learning} With the development of various loss functions, such as the multi-similarity loss \cite{wang2019multi}, and their use in face recognition \cite{kim2022adaface, deng2019arcface}, there has been interest in using deep metric learning-based losses for other recognition tasks such as fingerprint matching. The integration of AdaCos Loss with Contrastive Loss by Jawade et al \cite{wifs_bhavin} was shown to enhance the ability of deep convolutional networks to learn from minutiae features, improving contactless fingerprint recognition. Takahashi et al \cite{takahashi2020fingerprint} also utilize AdaCos in their CNN-based system to learn texture, minutiae, and frequency features from fingerprints.

\vspace{-1.0em}
\section{Method}
\label{sec:pagestyle}

\begin{figure*}[t!]
\centering
    \hspace{6em}
     \includegraphics[scale=0.7]{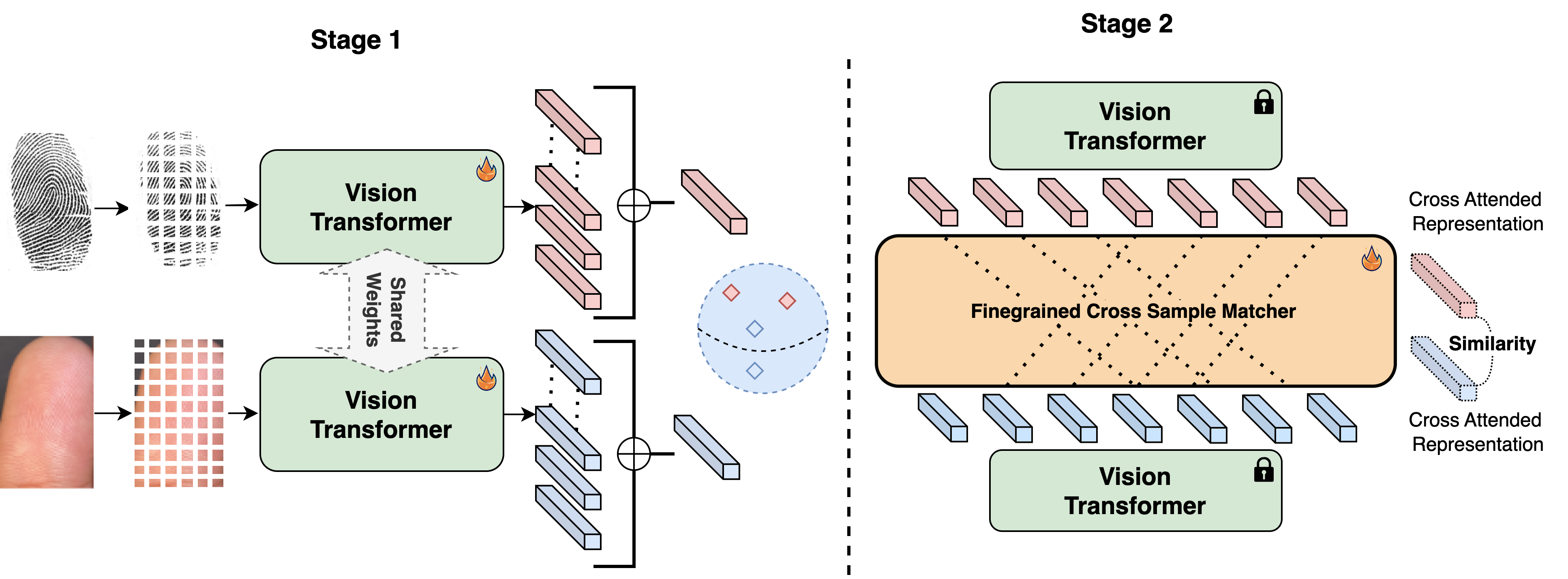}
    \caption{Illustration depicting our proposed transformer-based model architecture involving global features, fine-grained features, and domain features.}
    \label{fig:featureextractor}
\vspace{-1.5em}
\end{figure*}

\subsection{Problem Statement}
Let $D$ be a dataset consisting of contactless and contact-based fingerprint images from multiple subjects, represented as $\{V_0^s, V_1^s,..., V_n^s\}$ and $\{Q_0^s, Q_1^s,..., Q_n^s\}$, where $V_i^s$ denotes the $i^{th}$ contactless fingerprint image of subject $s$, and $Q_i^s$ denotes the $i^{th}$ contact-based fingerprint image of the same subject. The task is to learn a model $\theta$ that projects both $V_i^s$ and $Q_i^s$ into a common latent space such that the distance between the latent representations of fingerprints belonging to the same subject is minimized, while the distance between representations of different subjects is maximized. 

Let $\{v_0^s, v_1^s,..., v_n^s\}$ represent the low-dimensional contactless features, and $\{q_0^s, q_1^s,..., q_n^s\}$ represent the low-dimensional contact-based features. Formally, the model $\theta$ learns the following projection for both fingerprint types \(v_i^s, q_j^s = \theta(V_i^s), \theta(Q_i^s)\), where $v_i^s$ and $q_j^s$ are the feature vectors extracted from the contactless and contact-based fingerprints, respectively. The similarity between these features is computed using the cosine similarity between the normalized feature vectors \(d(v_i^s, q_j^s) = \frac{v_i^s \cdot q_j^s}{|v_i^s| |q_j^s|}\), where $\cdot$ represents the dot product between vectors, and $|v_i^s|$ and $|q_j^s|$ denote the magnitude of the feature vectors $v_i^s$ and $q_j^s$. The goal is to maximize $d(v_i^s, q_j^s)$ for matching pairs and minimize it for non-matching pairs.

\vspace{-1.0em}
\subsection{Architecture}
The proposed architecture is a two stage network. The first stage performs global feature extraction and the second stage performs localized cross sample alignment and matching for score computation. Below we provide details about the two stages:

\vspace{-1.0em}
\subsubsection{Stage 1: Global Feature Extraction}
In Stage 1, the goal is to extract global spatial features from both contactless and contact-based fingerprint images. We employ a Vision Transformer (ViT) \cite{vit} as the backbone feature extractor. Given an input fingerprint image $V_i^s$, we divide it into $T$ non-overlapping patches. Each patch is then projected into a $d$-dimensional embedding space using a linear embedding layer. Formally, for the $i^{th}$ contactless fingerprint image of subject $s$, the token embeddings are \({f_0^i, f_1^i, ..., f_{T-1}^i} = \theta(V_i^s),\) where $f_k^i$ represents the token embedding of the $k^{th}$ patch of the $i^{th}$ fingerprint image, and $\theta$ denotes the Vision Transformer. The transformer uses self-attention mechanisms to capture long-range dependencies between different regions of the fingerprint, which is crucial for representing global fingerprint patterns.

To obtain pooled global representation, we apply Global Average Pooling (GAP) over the set of token embeddings, followed by a linear projection and ReLU activation:
\[P^i = \frac{1}{T} \sum_{k=0}^{T-1} f_k^i, \quad v^i = \text{MLP}(P^i)\]

where $P^i$ is the pooled representation of the $i^{th}$ fingerprint image, and $v^i$ is the global representation after passing $P^i$ through a Multilayer Perceptron (MLP) with ReLU activation. This global representation $v^i$ is used as the input to subsequent stage, where local feature alignment is performed to address the fine-grained details that are lost in the global representation.

\subsubsection{Stage 2: Fine-grained Cross-Sample Matching}
In Stage 2, the goal is to achieve precise local alignment between contactless and contact-based fingerprint representations. Given that certain fine-grained details are lost in global feature extraction, we refine the matching process by leveraging token-level embeddings. 

Let $F_V^i = \{f_0^i, f_1^i, ..., f_{T-1}^i\}$ represent the token embeddings for the $i^{th}$ contactless fingerprint image $V_i^s$, and $F_Q^j = \{f_0^j, f_1^j, ..., f_{T-1}^j\}$ denote the token embeddings for the $j^{th}$ contact-based fingerprint image $Q_j^s$, both extracted from Stage 1. These token-level representations are then concatenated and passed through a cross-attention transformer module.

The transformer performs cross-attention between the token sets $F_V^i$ and $F_Q^j$, generating attended token representations \(F_V'^i, F_Q'^j = \text{CrossAttention}(F_V^i, F_Q^j)\).

Next, we apply global average pooling (GAP) to the attended token sets to obtain the final refined feature representations $v'^i$ and $q'^j$ for the contactless and contact-based fingerprints:
\[
v'^i = \frac{1}{T} \sum_{k=0}^{T-1} f_V'^k, \quad q'^j = \frac{1}{T} \sum_{k=0}^{T-1} f_Q'^k
\]

Finally, the cosine similarity between these pooled and aligned feature vectors is computed to measure the similarity between the contactless and contact-based fingerprints:
\[d(v'^i, q'^j) = \frac{v'^i \cdot q'^j}{|v'^i| |q'^j|}\]
This fine-grained matching ensures that local patterns are aligned effectively, improving matching accuracy between contactless and contact-based fingerprints.

\subsection{Optimization}
The primary training objective for both the stages is to enforce high-intra-class similarities and low-inter-class similarities. To do this, we utilize the multi-similarity loss \cite{wang2019multi}.

Using the features $v$ and the ground truth identity labels in the batch $B$, following \cite{wang2019multi}, we select hard positive and hard negative samples in the batch using the feature similarity matrix. Let $\text{S}_{ij}$ be the cosine similarity between two global features $v_i$ and $v_j$. 
We exponentiate the similarities while scaling them using a positive scale $(\alpha_{pos})$ and a negative scale $(\alpha_{neg})$, given by:
\begin{align*}
P = e^{-\alpha_{pos} \cdot (\text{S}_{ij} - \tau)} \ \forall \ \text{S}_{ij} \in B^+, \\ 
N = e^{\alpha_{neg} \cdot (\text{S}_{ij} - \tau)} \ \forall \ \text{S}_{ij}  \notin \ B^+
\end{align*}

where, $\tau$ represents the threshold. Next, we sample the hard negatives by selecting the negative samples that are closer than the farthest positive sample within a given margin. A similar approach is used to compute the hard positives. This is given by:
\begin{align*}
\text{P}_{hard} &= \sum \begin{cases} 
P_{ij}, & \text{if } P - \text{margin} < \text{max(N}_{i}\text{)} \\
0, & \text{otherwise} 
\end{cases} \\
\text{N}_{hard} &= \sum \begin{cases} 
N_{ij}, & \text{if } N + \text{margin} > \text{min(P}_{i}\text{)} \\
0, & \text{otherwise} 
\end{cases}
\end{align*}

\noindent Next, we perform scaled log-sum \cite{wang2019multi} to compute the positive and negative loss.
\begin{align*}
\text{P}_{loss} = \frac{\sum \log(1 + \text{P}_{hard})}{\alpha_{pos}}, \\
\text{N}_{loss} = \frac{\sum \log(1 + \text{N}_{hard})}{\alpha_{neg}}
\end{align*}
\[
\mathcal{L} = \text{P}_{loss} + \text{N}_{loss}     
\]

We compute the above loss for three similarity matrices (i) contactless to contactless similarity, (ii) contactless to contact-based similarity and (iii) contact-based to contact-based similarity. The final loss is given by \(\mathcal{L}_{G} = \mathcal{L}_{cl2cl} + \mathcal{L}_{cl2cb} + \mathcal{L}_{cb2cb} \).

\vspace{-0.5em}
\section{Experiments}
\label{sec:typestyle}

\subsection{Implementation Details}
In Stage 1 of our experiments, we trained our model using a combined training dataset. For the proposed loss function, we set the positive scale $(\alpha_{pos})$ to 2.0 and the negative scale $(\alpha_{neg})$ to 40.0. The margin for mining hard positives and hard negatives and the similarity threshold ($\tau$) were set to 0.7 and 0.5, respectively. The initial learning rate was set to $10^{- 5}$ with a decay factor of 0.3 applied at specific epochs, determined through experimentation. During fine-tuning in Stage 1, the learning rate was reduced to $5*10^{- 6}$, with a decay factor of 0.6 at designated epochs, while all other hyperparameters remained unchanged. Training was conducted with a batch size of 60 for 50 epochs on a single Nvidia A6000 GPU.

In Stage 2, we trained the model on the combined training dataset with a learning rate of $10^{- 5}$ and a batch size of 30. Both the margin and threshold ($\tau$) were set to 0.5. During fine-tuning on the HKPolyU dataset \cite{polyudataset}, the margin and threshold were reverted to the original Stage 1 configuration, with the learning rate maintained at $10^{- 5}$.

For all experiments, we utilized the AdamW optimizer with a weight decay of $10^{- 6}$.
\subsection{Datasets}
\subsubsection{Training Datasets}
Ridgeformer was trained on a comprehensive corpus that integrates four publicly available fingerprint datasets: the HKPolyU Contactless 2D to Contact-based 2D Fingerprint Images Database (HKPolyU data) \cite{polyudataset}, the IIITD SmartPhone Fingerphoto Database V1 (ISPFDv1)\cite{ispfdv1}, the IIITD SmartPhone Finger-Selfie Database V2 (ISPFDv2)\cite{ispfdv2}, and the Ridgebase dataset\cite{jawade2022ridgebase}.

The HKPolyU dataset \cite{polyudataset} was developed by researchers at Hong Kong Polytechnic University to facilitate the study of contactless-to-contact fingerprint matching. It includes 1,920 contactless fingerprints paired with corresponding 2D contact fingerprints, collected from 160 individuals. The RidgeBase benchmark dataset \cite{jawade2022ridgebase} was designed to address various fingerprint matching scenarios, including Single Finger Matching (Distal-to-Distal matching), Four Finger Matching, and Set-based Distal Matching, for both contactless-to-contactless (CL2CL) and contactless-to-contact (CL2CB) verification and identification. The training subset comprises approximately 11,000 fingerprint images from 63 subjects, captured under diverse lighting conditions and backgrounds using two types of smartphone cameras and a conventional flatbed contact sensor. In this study, we focus exclusively on Task 1 of the dataset, which falls within the scope of our research, while Tasks 2 and 3 are not considered within the scope of our analysis. The ISPFDv1 \cite{ispfdv1} and ISPFDv2 \cite{ispfdv2} datasets were prepared by researchers at IIT Jodhpur. ISPFDv1 \cite{ispfdv1} contains around 4,000 contactless fingerprints and 1,000 corresponding contact fingerprints, while ISPFDv2 \cite{ispfdv2} includes a significantly larger collection with approximately 16,800 contactless and 2,400 contact fingerprints. Both datasets utilized a random 50-50 split for training and testing through three rounds of validation.

\vspace{-0.5em}
\subsubsection{Evaluation Dataset}
Ridgeformer was assessed on the testing split of the HKPolyU dataset \cite{polyudataset}, which includes 960 contactless fingerprint images and their corresponding 960 contact fingerprint images from 160 distinct subjects, all disjoint from the training data. Additionally, we evaluated Ridgeformer on the testing split of Ridgebase dataset \cite{jawade2022ridgebase}, comprising 2,999 contactless images and 200 corresponding contact images from 25 subjects.\footnote{ISPDFv2's\cite{ispfdv2} official evaluation protocol requires a "random" subject-disjoint 50\%-50\% train-test split with three rounds of validation. We observe that because of this random nature of train-test split, performance varies significantly across experiments. Given this non-reproducability of their evaluation protocol, we don't report results on ISPDFv2\cite{ispfdv2}}

\vspace{-1.0em}
\subsection{Segmentation}
As part of the preprocessing workflow, different approaches were utilized to segment fingerprints from the original datasets. In the HKPolyU dataset \cite{polyudataset}, the contactless images were already closely cropped around the fingerprints, whereas the contact images contained padding around the region of interest (ROI) and thus required additional cropping.

In the ISPFDv1 \cite{ispfdv1} and ISPFDv2 \cite{ispfdv2} datasets, the contactless finger selfies were characterized by fingers occupying approximately 50\% of the frame, with the remaining space filled with various background elements, resulting in a visually cluttered context. The orientation of the fingers and the number of visible distal phalanges varied across images. To isolate the fingers, we applied a preprocessing technique combining the Segment Anything Model (SAM) \cite{SAM} and CLIP \cite{CLIP} to achieve precise cropping of the fingers. Additionally, the contactless images in ISPFDv1 \cite{ispfdv1} were rotated to align with the orientation of the contact fingerprints.

\setlength{\tabcolsep}{4pt} 
\renewcommand{\arraystretch}{0.95} 

\begin{table}[t!]
\begin{center}
\caption{1:1 Verification (Evaluated on HKPolyU\cite{polyudataset} and Ridgebase\cite{jawade2022ridgebase} datasets under different settings)}
\vspace{-0.5em}
\centering
\begin{tabular}{c@{\hskip 4pt}c@{\hskip 4pt}c@{\hskip 4pt}c@{\hskip 4pt}c}
\hline
\multicolumn{5}{c}{\textbf{HKPolyU Contactless to Contact 2D Dataset\cite{polyudataset}}} \\
\hline
Method & Probe & Gal. & EER\% & TAR@FAR=.01 \\
\hline
Verifinger & CL & CB & 19.31 & 76.00 \\
RTPS+DCM\cite{polyudataset} & CL & CB & 14.33 & 50.50 \\
Multi-Siamese\cite{lin2018cnn} & CL & CB & 7.93 & 54.00 \\
MANet\cite{tan2021minutiae} & CL & CB & 4.13 & 88.50 \\
ML Fusion\cite{wifs_bhavin} & CL & CB & 4.07 & \textbf{94.40} \\
\textbf{Ridgeformer} & CL & CB & \textbf{2.83} & 89.34 \\
\hline
\multicolumn{5}{c}{\textbf{Ridgebase Benchmark dataset\cite{jawade2022ridgebase} (Task 1)}} \\
\hline
Method & Probe & Gal. & EER\% & TAR@FAR=.01 \\
\hline
Verifinger & CL & CB & 18.90 & 57.60 \\
\textbf{Ridgeformer} & CL & CB & \textbf{5.25} & \textbf{82.23} \\
AdaCos(CNN)\cite{wifs_bhavin} & CL & CL & 21.30 & 61.20 \\
Verifinger & CL & CL & 19.70 & 63.30 \\
\textbf{Ridgeformer} & CL & CL & \textbf{7.60} & \textbf{85.14} \\
\hline
\end{tabular}
\label{tab:verification}
\end{center}
\vspace{-1.0em}
\end{table}

\begin{table}[t!]
\begin{center}
\caption{1:N Identification (Evaluated on HKPolyU\cite{polyudataset} and Ridgebase\cite{jawade2022ridgebase} datasets under different settings)}
\vspace{-0.5em}
\begin{tabular}{ccccc}
\hline
\vspace{-0.9em}
\\
\multicolumn{5}{c}{\textbf{HKPolyU Contactless to Contact 2D Dataset \cite{polyudataset}}} \\
\hline
\vspace{-0.6em}
\\
Method & Probe & Gallery & R@1 & R@10 \\
\vspace{-0.8em}
\\
\hline
ML Fusion\cite{wifs_bhavin} & CL & CB & - & - \\
RTPS+DCM\cite{polyudataset} & CL & CB & 66.67 & 83.00 \\
Multi-Siamese\cite{lin2018cnn} & CL & CB & 64.59 & 91.00 \\
Verifinger & CL & CB & 80.73 & 91.00 \\
MANet\cite{tan2021minutiae} & CL & CB & 83.54 & 97.00 \\
\textbf{Ridgeformer (Ours)} & CL & CB & \textbf{87.40} & \textbf{98.23} \\
\hline
\vspace{-0.9em}
\\
\multicolumn{5}{c}{\textbf{Ridgebase Benchmark dataset\cite{jawade2022ridgebase} (Task 1)}} \\
\hline
\vspace{-0.6em}
\\
Method & Probe & Gallery & R@1 & R@10 \\
\vspace{-0.8em}
\\
\hline
Verfinger & CL & CB & \textbf{72.50} & 89.20 \\
\textbf{Ridgeformer (Ours)} & CL & CB & 69.90 & \textbf{92.64} \\
Verfinger & CL & CL & 85.20 & 91.40 \\
AdaCos(CNN)\cite{wifs_bhavin} & CL & CL & 81.90 & 89.50 \\
\textbf{Ridgeformer (Ours)} & CL & CL & \textbf{100.00} & \textbf{100.00} \\
\hline
\end{tabular}
\label{tab:identification}
\end{center}
\vspace{-1.0em}
\end{table}

\begin{table}[t!]
\caption{Ablation Study (Evaluated on HKPolyU dataset \cite{polyudataset})}
\vspace{-0.5em}
\centering
\begin{tabular}{ccccc}
\hline
\vspace{-0.9em}
\\
Stage 1 & Stage 2 & Fine-tuned & EER\% & TAR\%@FAR=0.01 \\
\hline
\vspace{-0.4em}
\\
\centering \checkmark & - & - & 3.74 & 84.16 \\
\centering \checkmark & \checkmark  & - & 3.04 & 86.16 \\
\centering \checkmark & \checkmark  & \checkmark  & \textbf{2.83} & \textbf{89.34} \\
\hline
\end{tabular}
\label{tab:ablation}
\vspace{-1.0em}
\end{table}

For the Ridgebase benchmark dataset \cite{jawade2022ridgebase}, the contactless fingerprint images were similarly rotated to match the orientation of the contact fingerprints.

\subsection{Results and Discussion}
Table \ref{tab:verification} presents the results of 1:1 verification for both contactless-to-contact and contactless-to-contactless fingerprint matching. Ridgeformer outperforms all previously benchmarked models on the HKPolyU dataset\cite{polyudataset}, achieving an EER of less than 3\%. Additionally, when evaluated on the Ridgebase dataset\cite{jawade2022ridgebase}, Ridgeformer demonstrates significant improvements in both EER and TAR(\%)@FAR$=10^{-2}$, for both contactless-to-contactless and contactless-to-contact matching conditions. Ridgeformer consistently exceeds the performance of COTS and AdaCos(CNN)\cite{wifs_bhavin} by a considerable margin.
For 1:N Identification performance, Table \ref{tab:identification} shows that Ridgeformer achieved approximately a 4\% improvement in Rank-1 Recall compared to the best-performing model on the HKPolyU dataset\cite{polyudataset}, significantly enhancing identification performance. On the Ridgebase dataset\cite{jawade2022ridgebase}, Ridgeformer demonstrated competitive Rank-1 Recall performance in contactless-to-contact matching and exceeded expectations in contactless-to-contactless matching.

Our ablation study, presented in Table \ref{tab:ablation}, evaluates the impact of various components of our pipeline on performance using the HKPolyU dataset\cite{polyudataset}. The results indicate that employing only Stage 1 significantly reduced the EER, outperforming other methods listed in Table \ref{tab:verification}. Incorporating the Stage 2 architecture led to an additional decrease in EER and approximately a 2\% improvement in TAR(\%)@FAR$=10^{-2}$. Further fine-tuning the entire model on the HKPolyU dataset\cite{polyudataset} resulted in a further 0.5\% reduction in EER and a notable 3\% increase in TAR(\%)@FAR$=10^{-2}$. Overall, fine-tuning on the HKPolyU dataset\cite{polyudataset} provided a substantial enhancement of about 6\% in TAR(\%)@FAR$=10^{-2}$ and a 1\% decrease in EER.

\vspace{-1.0em}
\section{Conclusion}
\label{sec:majhead}

This paper introduces a novel framework for both contactless-to-contact and contactless-to-contactless fingerprint matching. The framework harnesses the feature extraction capabilities of vision transformers to derive learned embeddings from fingerprint images and applies cross-attention mechanisms to enhance matching performance. A unique deep metric learning loss function, which incorporates both local and global feature-based similarities, is employed to enforce accurate global and fine-grained representations of fingerprint images. Extensive evaluation on the HKPolyU \cite{polyudataset} and RidgeBase \cite{jawade2022ridgebase} datasets demonstrates that our approach effectively learns robust fingerprint representations, achieving notable performance improvements in both contactless-to-contact and contactless-to-contactless scenarios.

\bibliographystyle{IEEE}
\bibliography{Template}

\end{document}